\documentclass[runningheads]{llncs}
\usepackage{graphicx}
\begin{document}

\begin{center}
    \vspace*{1cm}
    \vspace{2cm}
    
    {\Large \textbf{Transfer learning for time series classification using synthetic data generation}}
    
    \vspace{1.5cm}
    
    Yarden Rotem ,\qquad  Nathaniel Shimoni,\qquad Lior Rokach,\qquad Bracha Shapira \\
    \textit{Software and Information Systems Engineering} \\
    \textit{Ben-Gurion University of the Negev}\\
    Be'er Sheva, Israel \\
    Email: \{rotemyar, nathanie, liorrk, bshapira\}@post.bgu.ac.il

    \vspace{1cm}
    
    In this paper\footnote{This preprint has not undergone peer review or any post-submission improvement or corrections. The Version of Record of this contribution is published in LNCS 13301, CSCML 2022, and is available online at \url{https://link.springer.com/chapter/10.1007/978-3-031-07689-3\_18}}, we propose an innovative Transfer learning for Time series classification method. Instead of using an existing dataset from the UCR archive as the source dataset, we generated a 15,000,000 synthetic univariate time series dataset that was created using our unique synthetic time series generator algorithm which can generate data with diverse patterns and angles and different sequence lengths. Furthermore, instead of using classification tasks provided by the UCR archive as the source task as previous studies did,we used our own 55 regression tasks as the source tasks, which produced better results than selecting classification tasks from the UCR archive.

  \vfill
\end{center}
\section{Abstract}
Both time series classification and transfer learning have increasingly been the focus of research in recent years. However, only a limited number of studies have combined time series classification with transfer learning.\\
Time series classification (TSC) is the task of training a classifier to map a given time series input to a probability distribution over the possible class values. Typically, transfer learning (TL) algorithms learn from a source dataset and task and then apply the knowledge gained to another target dataset and task.\\
TL has received considerable attention in the domains of computer vision and natural language processing, but less research attention has been devoted to the task of TSC, which is lacking a state-of-the-art pretrained model that can serve as a good starting point for new TSC tasks. 
All previous research in the domain of TL for TSC 
relied on existing datasets from the UCR archive, the largest publicly available TSC benchmark, in order to choose the optimal source dataset and task; there are some limitations to this approach, however. First, 
 searching for the optimal source dataset in the UCR archive can be time- and resource-consuming. Second, there is no guarantee that the optimal source dataset exists in the UCR archive. \\
In this paper, we propose an innovative TL for TSC method which addresses the limitations mentioned above. Instead of using an existing dataset from the UCR archive as the source dataset, we generated a 15,000,000 synthetic univariate time series dataset 
that was created using our unique synthetic time series generator algorithm which can generate data with diverse patterns and angles and different sequence lengths. Furthermore, instead of using classification tasks provided by the UCR archive as the source task as previous studies did,
we used our own 55 regression tasks as the source tasks, which produced better results than selecting classification tasks from the UCR archive. \\
With our unique source dataset and tasks, we pretrained a CNN (convolutional neural network) model and using 85 TSC datasets from the UCR archive to serve as target dataset, we performed an extensive evaluation of our method. We also reduced the training set of each dataset to only 10\% training data in order to emphasize the benefits of using TL for TSC when there is insufficient labeled data.\\
Our experimental results show that (1)
on datasets with seasonal characteristics, our method outperforms all other TSC methods (both TL and non-TL methods) on 17 of the 34 seasonal datasets in the UCR archive, whereas the second-best methods outperform on only seven of the 34 seasonal datasets; and
(2) the use of our method improves the test set's accuracy while reducing training time by 85\%, without compromising performance.
We published the code for the entire method which includes a synthetic time series data and regression task generator algorithm and a pretraining and fine-tuning process. We also published the 15,000,000 sample synthetic dataset and the pretrained CNN model.

\section{Introduction}
\textbf{Transfer learning (TL)} is a machine learning (ML) technique that tries to utilize knowledge learned from a source domain in a relevant target domain. The relevant knowledge is applied to the target domain in order to improve the performance of the prediction function of the target domain~\cite{zhao2017research}. The need for sufficient training data exists in most ML tasks, but obtaining labeled data  data can be expensive, time-consuming, or in some cases - infeasible. TL is a promising technique which can address this problem by transferring the knowledge across domains, preventing the need for labeled data in sparse domains~\cite{zhuang2019comprehensive}.   

TL has also shown to be effective at addressing some of the challenges with training a deep learning model which typically is time-consuming and requires high computational resources. Moreover, when lacking training data, ML models encounter the overfitting problem \cite{hawkins2004problem}.  

Transfer learning has also been widely used in computer vision, with state-of-the-art neural network (NN) models such as AlexNet~\cite{krizhevsky2012imagenet} and  ViT-G/14~\cite{zhai2021scaling}, which is the current leader in terms of top-1 accuracy on the ImageNet~\cite{deng2009imagenet} dataset. Evaluation of models pretrained on ImageNet show that accuracy is improved when using TL on new target datasets as opposed to training with the same architecture from scratch ~\cite{menegola2017knowledge}. TL has also been used effectively for natural language processing (NLP) tasks with pretrained models using word2vec and BERT models, and BERT's later versions were considered state of the art ~\cite{devlin2018bert}. Many studies used pretrained NLP models (such as BERT) to serve as a good starting point for new target datasets. However, for time series (TS) tasks, limited effort has been invested in developing a state-of-the-art, generic, and robust pretrained model that provides a good starting point for a new task.

A \textbf{time series} is a series of data samples in a time-based domain, which are typically sampled at a uniform time interval~\cite{karim2019multivariate}. There are two main types of TS: univariate time series (UTS) and multivariate time series (MTS)\cite{wang2016effective}.
An MTS is an $M$-dimensional TS where each data sample consists of $M$ real values, e.g., an MTS can be data acquired by measuring multiple climate sensors, such as temperature, humidity, and wind speed, once an hour; this is an $M=3$ MTS. A UTS is simply an MTS where $M=1$; a UTS can be data acquired by sampling the heartbeat of a patient every 10 seconds~\cite{IsmailFawaz2019inceptionTime}.
In this paper, we focus only on UTS data.
TS data is relevant for many domains, including the analysis of financial transactions~\cite{zhu2002statstream}, monitoring network traffic~\cite{papadimitriou2006optimal}, the analysis of time-based medical events ~\cite{keogh2001online}. In fact, TS data mining was mentioned as one of the top 10 data mining problems by Yang and Wu~\cite{yang200610}.\\
Time series data analysis is a highly focused research domain that has a number of different applications. The three main applications are: \textbf{time series classification (TSC)} - the task of training a classifier to map a given input to a probability over the possible class values (labels)~\cite{fawaz2019deep}, time series forecasting - the task of predicting future values of a given sequence using previous data~\cite{sagheer2019time}, and time series clustering - the task of dividing a set of TS data into groups, where similar TS samples are put in the same cluster~\cite{ferreira2016time}. In this work, we  focus only on time series classification.\\
\textbf{TL for TSC} has not been extensively studied, and a generic, robust, and scalable pretrained model that can serve as a good starting point for new datasets is needed, especially when there is insufficient labeled data. Due to the time-consuming process of collecting and labeling data, the availability of such a pretrained model is essential and would reduce the training time and cost, and in some cases, these models could lead to better overall results. \\\\
In this study, we propose an innovative, generic, scalable, and architecture-agnostic TL for TSC method based on (1) our new algorithm for generating synthetic data and (2) 55 corresponding  regression tasks. Our method can be applied to any deep learning CNN-based architecture.
As opposed to previously proposed TL for TSC methods, our model only needs to be pretrained once, and there is no need to search for the optimal source dataset for every new target dataset. Using our unique algorithm, 15,000,000 synthetic samples of UTS data with various angles, sequence lengths, and patterns were used to pretrain our CNN (convolutional neural network) model.\\
Using 85 datasets from the UCR archive as target datasets, we perform a comprehensive evaluation of our method.
For datasets with seasonal characteristics, when the amount of training data was reduced to 10\%, our method outperforms all other TSC methods (both TL and non-TL methods) on 17 of the 34 seasonal datasets in the UCR archive, whereas the second-best methods outperform on only seven of the 34 seasonal datasets.\\
Additionally, using
our method improves the test set's accuracy while reducing training time by 85\%, without compromising performance. 
We thus believe that our method can serve as a good starting point for any new target dataset.\\  
The contributions of this paper are as follows:
\begin{enumerate}
    \item
    \textbf{\emph{Synthetic UTS data and regression task generator algorithm: }}In this paper, we contribute a new architecture-agnostic TL for TSC method. Unlike previously proposed TL for TSC methods which use an existing source dataset and classification task from the UCR archive \cite{dau2019ucr}, we propose a new algorithm which generates synthetic UTS data and creates 55 corresponding regression tasks which can be used as a source dataset and task. \\
    Using existing datasets from the UCR archive as the source dataset has some limitations that our synthetic data overcomes. First, given a target dataset, datasets from the UCR archive may not always be similar or generic enough to serve as a good source dataset. Since our synthetic 15,000,000 sample dataset has a wide variety of patterns, angles, and sequence lengths, it could be a more generic source dataset and therefore be a better fit. A second limitation is that using UCR datasets is not scalable: each update to the UCR archive requires that TL for TSC methods perform a new pretraining procedure to incorporate the new datasets. Since our method relies on the synthetic 15,000,000 sample dataset as a source dataset, no additional pretraining is necessary. Finally, searching for the optimal source dataset from the UCR archive can be time- and resource- consuming. Since we do not use datasets from the UCR but instead use our synthetic dataset, no such search is needed. \\
    In this paper, we demonstrate the superiority of a dataset consisting of synthetic data over existing datasets from the UCR archive, by addressing all of the above mentioned issues.
    \item
    \textbf{\emph{Code contribution: }}Our code\footnote{Code availble at: \url{https://github.com/YR234/TL-for-TSC}}  includes the following: 
    \begin{itemize}
    \item
    \textbf{\emph{UTS data and regression tasks generator: }}We created an algorithm to generate synthetic UTS data with a wide range of UTS patterns, angles, and sequence lengths that can serve as a source dataset.
    \item
     \textbf{\emph{Complete framework: }}a comprehensive easy-to-use framework that covers data and regression task generation through fine-tuning the pretrained model on a new target dataset and task.
     \end{itemize} 
     \item  We publish both the synthetic dataset with 15,000,000 UTS samples and the pretrained CNN model with the $CTN$ architecture that was pretrained on that dataset, making them available for use by researchers and the entire ML community.
\end{enumerate}
The remainder of the paper is structured as follows: In section~\ref{background and related work}, we provide the necessary background and introduce related work on TSC and TL for TSC methods. Following this, in section~\ref{method}, we describe our method, from data generation through fine-tuning the pretrained model on a new target dataset and task.
In section~\ref{experiment}, we describe the experimental setup, while section~\ref{results} presents our results. Finally, in section~\ref{conc}, we present our conclusions and plans for future work.

\section{Background and related work}
\label{background and related work}
In this section, we first discuss on related work regarding TSC and TL for TSC methods, and we highlight the differences between those methods and ours.

\subsection{TSC related work}
\label{TSC-related}
In this subsection, we discuss previously proposed TSC methods. \\
\textbf{MultiRocket}~\cite{tan2021multirocket} is a TSC method that achieved SOTA (state-of-the-art) results on the entire UCR archive~\cite{dau2019ucr} at a rate orders of magnitude faster than any other competing method.\\
MultiRocket is, in practice, a single-layer convolutional neural network, 
where the transformed features from the convolutional kernels form the input for a linear classifier. \\
MultiRocket uses as many as 10,000 convolutional kernels with a wide range of length, padding, dilation, and random weights. After the kernels are generated, each kernel is  applied to each input time series, resulting in a feature map. MultiRocket then computes a set of features from the feature map that includes PPV (portion of positive values) plus a randomly selected features from
a set of five candidate features. These features serve as the input for a linear classifier, such as a ridge regression classifier or logistic regression.\\ 
MultiRocket does not use a nonlinear function or have any hidden layers, thus allowing it to be orders of magnitude faster than any other method.\\
\textbf{OS-CNN}~\cite{tang2020rethinking} is a TSC method that uses omni-scale (OS) blocks, which does not need to tune the feature extraction scales. Usually, a core challenge of a CNN is to determine the proper scales of feature extraction. This method uses OS blocks which are made up of OS layers that can be configured automatically from the input size based on a list of kernel sizes; by stacking those layers, this method can achieve full receptive field coverage of the total length of the input (sequence length)~\cite{he2016deep}.\\
\textbf{InceptionTime}~\cite{fawaz2020inceptiontime} is a TSC method that uses an ensemble of five deep CNN models, which was inspired by the Inception-V4~\cite{szegedy2017inception} architecture. This architecture includes several techniques commonly used when constructing a CNN model, such as residual block with shortcut connections~\cite{he2016deep} and inception modules~\cite{szegedy2017inception}. Each of the five models is given equal weight in the final prediction decision. \\
Because of our TL-based approach, our method differs entirely from the TSC methods mentioned above. In the absence of sufficient labeled data, TL techniques are useful. In this paper, we leverage this by reducing the amount of labeled training data to 10\%. Our experimental results indicate that when it comes to seasonal datasets, our method outperforms all other methods, and with all datasets (both seasonal and non-seasonal) our method is only second to MultiRocket, however the difference in the performance of the two methods was not shown to be significant when the Nemenyi statistical test was performed.

\begin{figure}[tbh!]
\advance\leftskip-0.2cm
\includegraphics[width=1.1\columnwidth]{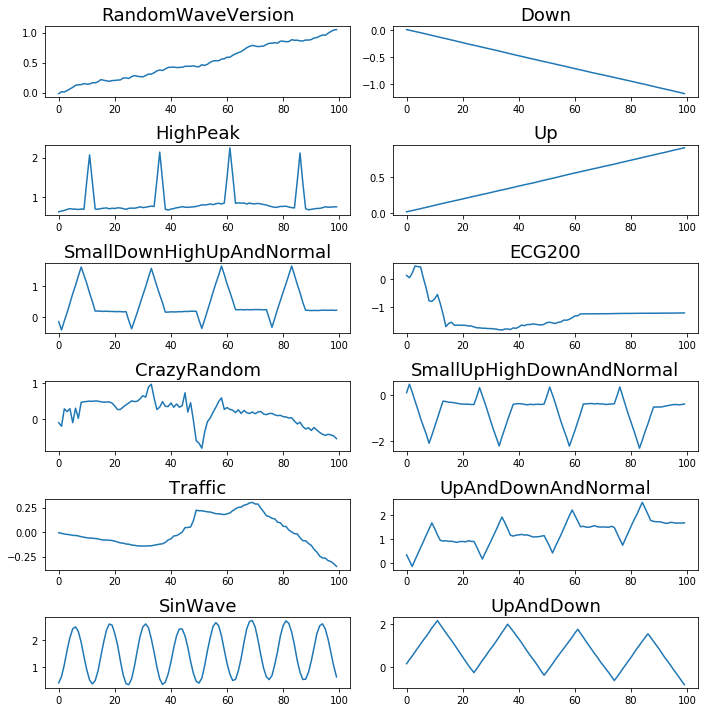}
\caption{The 12 UTS patterns generated in our work.}
\label{waves}
\end{figure}

\subsection{TL for TSC related work}
\label{TL-TSC-related}
The use of TL for TSC has been proposed in a number of studies. In this subsection, we discuss the existing TL for TSC methods and how our TL for TSC method differs from these methods.\\
An overview of the general TL for TSC process is presented in Fig.~\ref{method_overview}. This process consists of the following five steps: First, a source dataset is selected. Second, a source task is selected. In step 3, the model's architecture is chosen. In step 4, the model chosen in step 3 is pretrained on the source dataset and task selected respectively in steps 1 and 2. The final step consists of fine-tuning the pretrained model from step 4 on a new target dataset and task. \\
While all previous TL for TSC studies used existing datasets and classification tasks from the UCR archive as the source dataset and task for steps 1 and 2, in this paper, we generate synthetic data for the source dataset and use regression tasks instead of classification as the source task, and demonstrate how those two decisions can result in better generalization while eliminating the need for an exhaustive search for the best source dataset.\\
\textbf{Fawaz et al}~\cite{fawaz2018transfer} suggested using DTW (dynamic time warping), a technique for finding the optimal alignment between two given time series sequences~\cite{muller2007dynamic}, as a similarity measure for finding the most similar source dataset from the UCR archive. The source task is chosen according to the source dataset (provided by the UCR archive). \\
While our method may only differ from the method of Fawaz et al in terms of steps 1 and 2 of the TL for TSC process, our novel approach for creating the source dataset and task from synthetic data and regression tasks instead of using an existing dataset and classification task from the UCR archive addresses other issues that we will discuss later in the paper.\\
Our experimental results on the UCR archive showed that the method proposed by Fawaz et al performed positive transfer learning on $71/85$ datasets, however this approach has some disadvantages. \\
\textbf{Kashiparekh et al}~\cite{kashiparekh2019convtimenet}\label{convtimenet} suggested using a convolutional neural network (CNN) based architecture with a multi-head approach for training a given $S$ source dataset ($D_S$) and corresponding  $S$ classification tasks ($T_S$) from the UCR archive.\\
The CNN core architecture consists of convolutional layers followed by skip connections~\cite{he2016deep}, which make this architecture a deep one. However, instead of standard fully connected layers followed by a dense layer with the softmax activation function, the authors used $S$ fully connected layers and $S$ dense layers with the softmax activation function - one for each source dataset and task. \\
The authors randomly selected $S=24$ datasets from the UCR archive for training and validation, and the remaining 41 datasets were used as test sets (the authors used sequence lengths up to 512, and therefore not all 85 datasets of the UCR archive were evaluated).
\\ As noted earlier, none of these methods provides a real solution when it comes to real-world problems in the domain of TL for TSC. Since they are limited to the available datasets in the UCR archive, they may not always be able to find the optimal source dataset.
Not only that, when using the method proposed by Fawaz et al, one would have to perform an exhaustive search to find the most similar source dataset for a new target dataset and task.\\ 
In contrast to prior work, our method does not require an exhaustive search, and it is not restricted to datasets available in the UCR archives or any specific sequence length. Since it is based on diverse synthetic data that was generated by our new algorithm, it can be applied to a variety of new target datasets and tasks.

\newcommand{\SubItem}[1]{
    {\setlength\itemindent{15pt} \item[-] #1}
}

\section{Method}
\label{method}
\begin{figure*}[tbh!]
\centering
\hspace*{-5cm} 
\includegraphics[width=1.8\columnwidth]{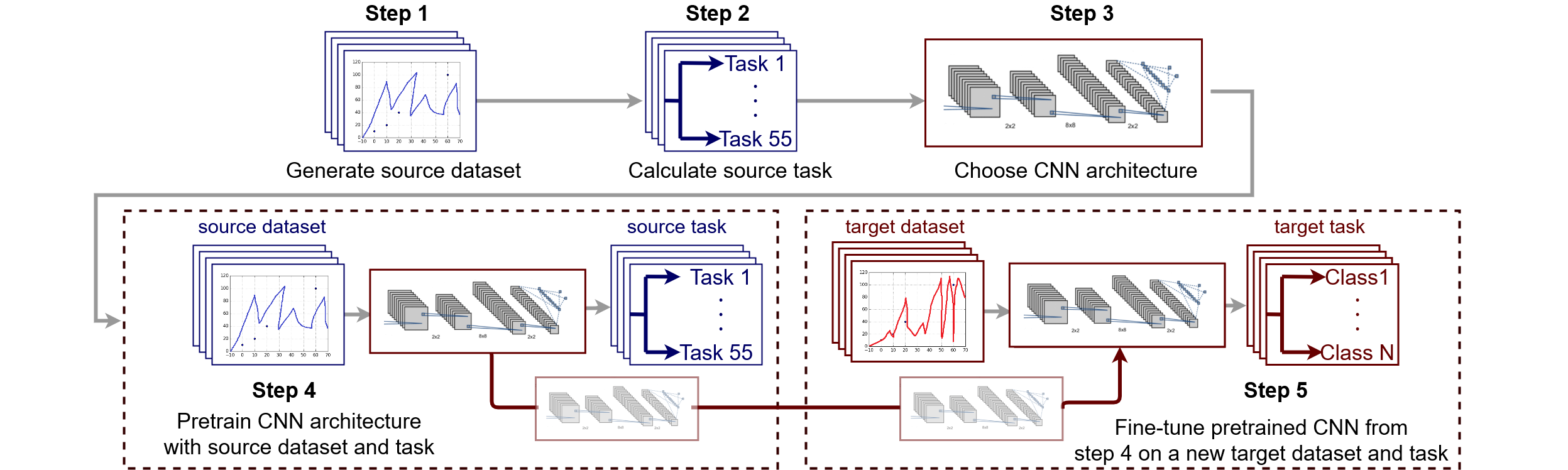}
\caption{\textbf{Method overview}: In step 1, we generate a a 15,000,000 sample UTS  source dataset using our Algorithm. After that, we calculate 55 regression tasks for each UTS in the source dataset to be our source tasks. In step 3, we select the CNN architecture (we chose to use $CTN$). Then in step 4, we train the CNN with the source dataset and task. Finally, in step 5, we fine-tune the pretrained CNN model on a new target dataset and task.}
\label{method_overview}
\end{figure*}

In this section,We will discuss on our five-step TL for TSC method (see Fig.~\ref{method_overview}). The first two steps describe the process of generating the source dataset and regression source task. We then describe steps 3-5 where we select and pretrain the CNN architecture and fine-tune the pretrained model on a new target dataset and task.
\subsection{Data generation - source dataset}
\label{met:input}
We created a synthetic UTS data generator algorithm; using this algorithm, we created a 15,000,000
sample dataset that contains a wide range of UTS with different segment patterns, angles, sequence lengths. This dataset will serve as our source dataset $D_S$.\\
In our study, we generated only 12 UTS patterns.
However, using our algorithm, many more patterns can be generated.

\subsection{Data generation - source tasks}
\label{met:target}
Upon generating the source dataset, we proceed to the source task.
Because our target task is TSC (classification), it would be natural to use classification as our source task, however when comparing classification and
regression as source tasks,
we found that regression achieves more accurate results,
and therefore it was chosen as our source task.
\begin{enumerate}
\item The 55 tasks are:
\renewcommand{\labelenumii}{\theenumii}
\renewcommand{\theenumii}{\arabic{enumii}.}
\begin{enumerate}
\item \label{max} \textbf{Maximum (Task 1):} Given an input UTS, the purpose of the task is to accurately predict the maximum value of the UTS.
\item \textbf{Minimum (Task 2):} Given an input UTS, the purpose of the task is to accurately predict the minimum value of the UTS.
\item \textbf{STD (Task 3):} Given an input UTS, the purpose of the task is to accurately predict the STD (standard deviation) value of the UTS.
\item \textbf{Peaks (Task 4):} Given an input UTS, the purpose of the task is to accurately predict the number of high and low peaks.
\item \label{mean} \textbf{Cross median (Task 5):} Given an input UTS, the purpose of the task is to accurately predict the number of times the UTS crosses the median value from up to down and vice versa.
\item \textbf{10 splits (Tasks 6-55):} Given an input UTS, we first divide the UTS into 10 equal length segments. For each segment we calculate tasks 1-5 and concatenate them into a 50 value task (10 segments * 5 tasks).
\end{enumerate}
\end{enumerate}

\subsection{CNN model's architecture}
\label{met:archi} Our method is architecture-agnostic, meaning that any deep learning network with a convolutional layer based architecture (CNN) can be used. In our research we used the same CNN architecture as Kashiparekh et al~\cite{kashiparekh2019convtimenet} whose work showed it to be an effective architecture for TL. Just one change was made to their architecture; unlike the multi-head approach used by Kashiparekh et al., we used only one dense layer with the sotfmax activation function. This architecture will be denoted as $CTN$.
\subsection{CNN pretraining}
\label{met:training}
The next step of our method is pretraining the $CTN$ model on the source dataset $D_S$ with the source task $T_S$.\\
To create the training and validation sets, we randomly divided the source dataset $D_S$ into an 80\%-20\% split.
We pretrained the $CTN$ model for 100 epochs with a batch size of 128, while performing early stopping on the validation loss. We used the $Adam$ \cite{kingma2014adam} optimizer and $MSE$ (mean square error) as the loss function. \\ 
The pretraining process was lengthy, taking almost 10 days on a single GPU processor, however this process only needs to happen once. Once the $CTN$ model has been pretrained, we save the weights of the model's core layers. These weights are used later to fine-tune new target datasets and tasks.
\subsection{Fine-tuning a new target dataset}
\label{met:fine}
The final step of our method is to fine-tune the pretrained $CTN$ model on a new target dataset $D_T$ with a new target task $T_T$. \\
We start by initializing the $CTN$ model's core layers with the pretrained weights that were saved in the last step. Once initialization is complete, newly added fully connected layers can be adjusted so that they better fit the new target dataset and task.

\section{Experimental setup}
\label{experiment}
In this section, we describe our experimental setup, including the datasets, preprocessing, and settings used, as well as the methods we compare our method to.

\subsection{Datasets}
\label{exp:UCR}
To evaluate our method, we used the UCR archive, which is the benchmark archive for TSC. In 2002, the archive contained 45 datasets; this increased to 85 datasets in 2015, and as of 2019, 128 datasets are available. All of the datasets contain UTS samples. The archive's datasets vary in terms of the time series domain covered; the domains include traffic, sound, sensors, motion, image, HAR (human activity recognition), financial, medical, and more. Furthermore, they are diverse in terms of the sequence length (ranging from 8-5,000), number of classes (2-60), number of training samples (12-139,000), and number of test samples (15-139,000). Due to running time considerations, we evaluated our results on just the 85 dataset version of the UCR archive and not on the most update version that includes 128 datasets. 
\subsection{Data preprocessing (reducing the labeled training data to 10\%)}
\label{exp:pre}
The original train-test split provided by the UCR archive was used. However, instead of using all of the training data, we reduced each dataset's training data to only 10\% of the original, while keeping the same class distribution, e.g., given a dataset of 100 training samples with 70 samples of class 1 and 30 samples of class 2, we reduced the training data to 10 samples, with seven class 1 samples and three class 2 samples. Therefore, the 70-30\% class distribution was maintained. \\
The following are some key points regarding the reduction process:
\begin{enumerate}
    \item The remaining 10\% of the training samples were chosen at random.
    \item The reduction process was only performed once.
    \item All of the test data samples were evaluated (the test data was not reduced).
\end{enumerate}
This reduction process was used to emphasize the importance of transfer learning, since when there is a lack of labeled data, the pretraining process (steps 1-4 of our method) is expected to provide a better start for learning a new target dataset and task than learning from scratch.
\subsection{Methods used for comparison}
\label{exp:methods}
We compare our method, which will now be denoted as $CTN\_our$, to the methods covered in the related work section: Fawaz et al~\cite{fawaz2018transfer} (denoted as $Fawaz$), Kashiparekh et al~\cite{kashiparekh2019convtimenet} (denoted as $ConvTime$), MultiRocket~\cite{tan2021multirocket},  OS-CNN~\cite{tang2020rethinking}, and  InceptionTime~\cite{IsmailFawaz2019inceptionTime}; we also examine the $CTN$ architecture without the pretraining phase (steps 1-4 in our method), which is denoted as $CTN\_S$ (no transfer learning was applied). We included $CTN\_S$, so we can examine our results in terms of positive and negative transfer learning and more. 

\subsection{Hyperparameters and other settings}
\label{exp:hyper}
For all deep learning methods (Fawaz et al, Kashiparekh et al, OS-CNN, Inception-Time), including ours, we trained each dataset with 2,000 epochs, using cross-entropy~\cite{de2005tutorial} as the loss function and Adam~\cite{kingma2014adam} as the optimizer. \\
For MultiRocket (a linear classifier), we used the default parameters provided by the authors.
\subsection{The evaluation process}
\label{exp:evaluation}
The evaluation process includes applying all of the methods on the datasets with the necessary data preprocessing and with the hyperparameters - all this was described at this section. \\
A summary of the results is provided in the next section.

\section{Results}
\label{results}
This section begins with a brief summary of the results. We then explore each aspect mentioned in the brief summary in more detail. \\
\subsection{Results appendices} 
\begin{figure}
\hspace*{-4.2cm} 
\includegraphics[width=1.75\columnwidth]{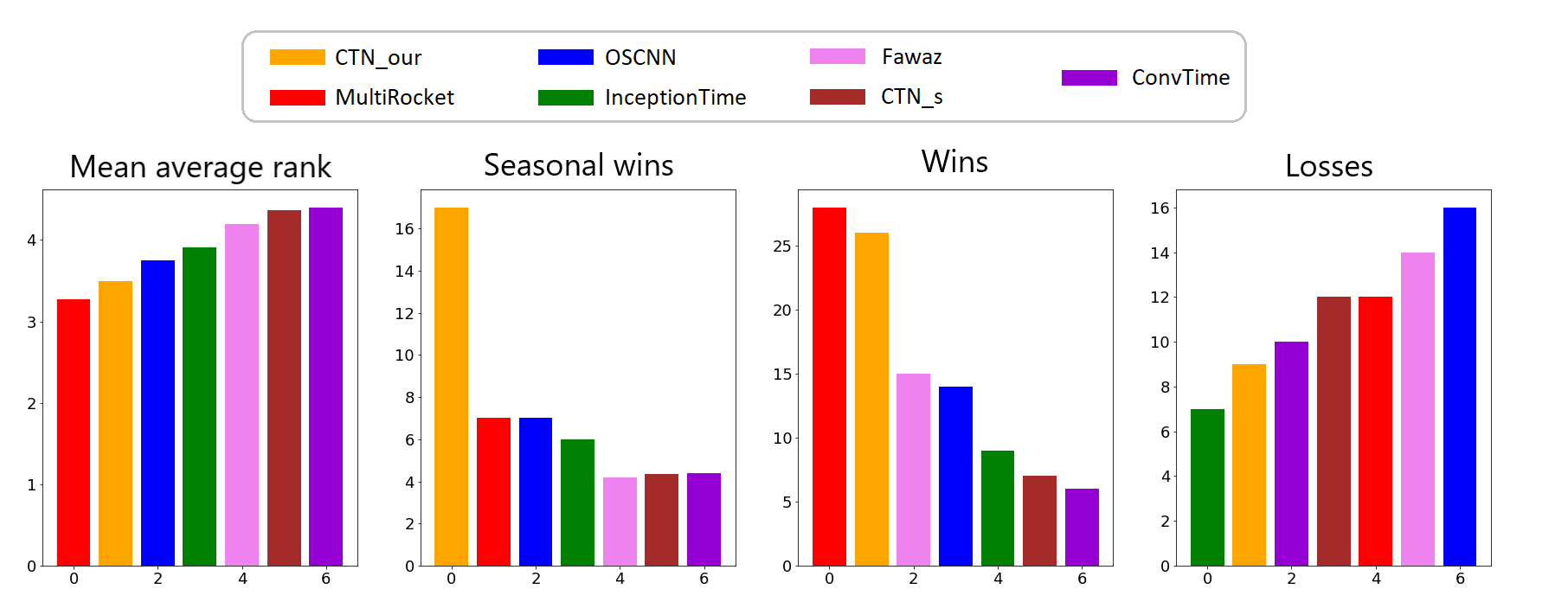}
\caption{Empirical results, from left to right: mean average rank, seasonal wins, wins, losses. Each method is associated with a colored bar.}
\label{results_all}
\end{figure} 

A summary of the results can be seen in Table~\ref{results_table} and a more visual representation can be found in Fig.~\ref{results_all}.
\begin{table}[]
\caption{Summary of the results in terms of the number of wins, number of losses, seasonal wins, and mean average rank}
\begin{tabular}{ccccc}
\hline
Method        & Wins                 & Losses              & Seasonal Wins         & Mean Avg Rank           \\ \hline
ConvTime      & 6                    & 10                  & 3                    & 4.4                     \\
CTN\_S        & 7                    & 12                  & 4                    & 4.365                   \\
Fawaz         & 15                   & 14                  & 4                    & 4.2                     \\
InceptionTime & 9                    & \textit{\textbf{7}} & 6                    & 3.906                   \\
OSCNN         & 14                   & 16                  & 7                    & 3.753                   \\
CTN\_our      & 26                   & 9                   & \textit{\textbf{17}} & 3.494                   \\
MultiRocket   & \textit{\textbf{28}} & 12                  & 7                    & \textit{\textbf{3.271}}

\end{tabular}
\label{results_table}
\end{table}

\subsection{Brief summary of the results}
\label{results:brief}
\begin{enumerate}
    \item Thirty-four of the 85 UCR archive datasets have seasonality characteristics. Of these datasets, our method outperforms all other examined methods on 17 datasets; the second best method only outperforms all other methods on seven datasets.
    
    \item Positive transfer learning occurs in all 2,000 epochs except the first seven epochs, and using our method can save 85\% of the training time while achieving the same results (see Fig.~\ref{positive_transfer_graph}).
    
    \item Our method obtains a mean average rank of 3.494, which is second only to MultiRocket with a mean average rank of 3.271 (the difference between the two values is not significant. In terms of the win/lose rate, our method obtains a 26/9 rate, while MultiRocket's rate is 28/12; MultiRocket has two more wins, but it also has three more loses than our proposed method. 
    
    \item As can been Fig.~\ref{results_all}, our method comes in at least second place in each case, something no other method achieved.
    
\end{enumerate}
\subsection{Seasonality evaluation}
\label{results:seasonality} We first evaluate the results in terms of seasonality. The autocorrelation values of a given UTS are in the range of $[-1,1]$. Generally, as the autocorrelation values approach zero there is no seasonality and vice versa. So, in this paper, we took the absolute value of the autocorrelation function, and the values will eventually be in the range of $[0,1]$. Higher absolute values of the autocorrelation function indicate strong seasonality and vice versa. \\
To empirically define datasets with seasonality we calculated a seasonality metric for each dataset, which is denoted as $SM$ (seasonality metric). \\

All datasets with $SM => 0.5$ will be considered as seasonal datasets.
The results show that 34 of of the 85 UCR archive datasets are seasonal datasets. Of these 34 datasets, $CTN\_our$ outperforms all other methods on 17 datasets. The second best methods are OSCNN and MultiRocket which outperform other methods on only seven of the 34 seasonal datasets. These results are presented in Table~\ref{results_table} and more visually in Fig.~\ref{results_all}.
These results indicate that our method is superior when it comes to seasonal datasets.

\subsection{Positive transfer learning}
\label{results:positive}
\begin{figure}
\centering
\includegraphics[width=1\columnwidth]{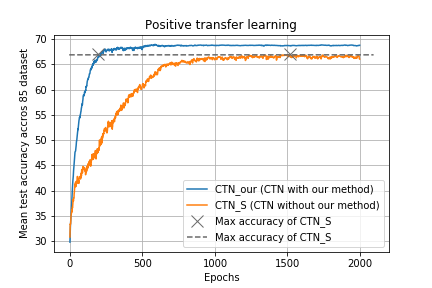}
\caption{Performance of the $CTN$ architecture on all of the test sets from the UCR archive after each epoch, with and without our method.}
\label{positive_transfer_graph}
\end{figure}

To demonstrate that transfer learning using our method can improve the performance of a given CNN architecture, we compered the following two methods: $CTN\_our$ and $CTN\_S$. The comparison was made while considered the following three aspects: positive transfer learning after each epoch, training time, and final accuracy. The  results can be seen in Fig.~\ref{positive_transfer_graph}.

A more detailed evaluation for each of the aspects mentioned above is provided below:
\begin{enumerate}
\item
    \textbf{\emph{Positive transfer learning: }}Except for the first seven epochs, $CTN\_our$ outperforms $CTN\_S$. Accordingly, we can conclude that our method's performance improves when training continues beyond the first few initial training epochs.
    
    \item
    \textbf{\emph{85\% Less training time: }}It took $CTN\_S$ 1,515 epochs to achieve its highest level of accuracy on all 85 datasets (indicated by the grey 'X' in Fig.~\ref{positive_transfer_graph}), whereas $CTN\_our$ attains the same level of  accuracy after only 193 epochs. In other words, using our method can save as much as 85\% of the training time and still achieve the same generalization.
    
     \item \textbf{\emph{Accuracy: }}When considering the average accuracy after 2,000 epochs across the 85 datasets from the UCR archive, $CTN\_our$ outperforms $CTN\_S$, as can be see in Fig.~\ref{positive_transfer_graph}.
     
\end{enumerate}

\subsection{Mean average rank and win/lose rate}
\label{results:mar}
In terms of the mean average rank, $CTN\_our$ achieved a score of \textbf{3.494}, which is only second to MultiRocket
with a score of 3.271; the Nemenyi statistical test indicated that there was no significant difference in the results of the two methods 

(see Table~\ref{results_table} which presents the mean average rank of all of the examined methods).\\
In terms of the win/lose rate, our method obtains a \textbf{26/9} win/lose rate. While MultiRocket achieves a 28/12 win/lose rate, winning two more times but losing three more times. InceptionTime has the fewest losses, with a win/lose rate of 9/7.\\
Our method comes in at least second place on every empirical measure (mean average rank, seasonal wins, win, lose), which no other method is capable of (the results of all of the examined methods can be seen in Table~\ref{results_table}).

\section{Conclusion and future work}
\label{conc}
In this paper, we introduced: \begin{enumerate}
    \item A novel architecture-agnostic TL for TSC method. In contrast to previous TL for TSC methods using existing UCR's datasets and tasks as the source dataset and task, in this paper, we introduce a new algorithm that generates UTS data and creates 55 corresponding regression tasks to be used as a source dataset and task.
    \item Open-source code for generating custom synthetic data, producing regression tasks, pretraining, and fine-tuning on a new target dataset and task. \\
    Our 15,000,000 sample synthetic dataset, the 55 regression tasks, and the pretrained model with the $CTN$ architecture are published for further use by the ML community.
\end{enumerate}
Our study shows that the use of our method can not only improve the performance of a given CNN architecture but also decreases training time by 85\%. \\
When it comes to seasonal datasets, our method outperforms all other existing TSC  methods.\\
For future work, we would like to do the following:
\begin{enumerate}
    \item Explore other regression tasks beside the 55 tasks we used. The new tasks could include Fourier transform, autocorrelation, etc.
    \item Examine a new architecture for transfer learning based on both CNN and LSTM layers which will be trained on our source dataset and task, as suggest by Wang et al \cite{wang2019time}.
    \item Expand our source dataset to include more patterns than the 12 patterns we generated.
    \item Extend our method so that it will be suitable for MTS (multivariate time series) datasets.
    \item Explore the use of a generative adversarial network (GAN) to automatically generate synthetic data that is more similar to a specific given dataset.
\end{enumerate}

\bibliographystyle{plain}
\bibliography{bibliography.bib}
\end{document}